\documentclass[conference]{IEEEtran}
\IEEEoverridecommandlockouts

\usepackage{cite}
\usepackage{amsmath,amssymb,amsfonts}
\usepackage{algorithmic}
\usepackage{graphicx}
\usepackage{epsfig}
\usepackage{epstopdf}
\usepackage{textcomp}
\usepackage{xcolor}
\usepackage{float}

\usepackage[
  letterpaper,
  left=0.75in,
  right=0.75in,
  top=0.75in,
  bottom=1.05in
]{geometry}

\def\BibTeX{{\rm B\kern-.05em{\sc i\kern-.025em b}\kern-.08em
    T\kern-.1667em\lower.7ex\hbox{E}\kern-.125emX}}

\begin{document}

\title{Adaptive Semantic Communication for Wireless Image Transmission Leveraging Mixture-of-Experts Mechanism}
\author{\IEEEauthorblockN{Haowen Wan$^{1,2}$, Qianqian Yang$^{2}$}
\IEEEauthorblockA{$^1$ Polytechnic Institute, Zhejiang University, Hangzhou, China}
\IEEEauthorblockA{$^2$ College of Information Science and Electronic Engineering, Zhejiang University, Hangzhou, China}
\IEEEauthorblockA{Email:\{haowenwan20, qianqianyang20\}@zju.edu.cn}
\thanks{This work is partly supported by the National Key R\&D Program of China under Grant No. 2025YFF0514602, by the National Natural Science Foundation of China (NSFC) under Grant Nos. 62293481, 62571487, by the Zhejiang Provincial Natural Science Foundation of China under Grant No. LZ25F010001.}
}

\maketitle
\thispagestyle{empty}
\pagestyle{empty}

\begin{abstract}
Deep learning based semantic communication has achieved significant progress in wireless image transmission, but most existing schemes rely on fixed models and thus lack robustness to diverse image contents and dynamic channel conditions. To improve adaptability, recent studies have developed adaptive semantic communication strategies that adjust transmission or model behavior according to either source content or channel state. More recently, MoE-based semantic communication has emerged as a sparse and efficient adaptive architecture, although existing designs still mainly rely on single-driven routing. To address this limitation, we propose a novel multi-stage end-to-end image semantic communication system for multi-input multi-output (MIMO) channels, built upon an adaptive MoE Swin Transformer block. Specifically, we introduce a dynamic expert gating mechanism that jointly evaluates both real-time CSI and the semantic content of input image patches to compute adaptive routing probabilities. By selectively activating only a specialized subset of experts based on this joint condition, our approach breaks the rigid coupling of traditional adaptive methods and overcomes the bottlenecks of single-driven routing. Simulation results indicate a significant improvement in reconstruction quality over existing methods while maintaining the transmission efficiency.

% Deep learning based semantic communication promises unprecedented transmission efficiency but faces critical challenges regarding high computational overhead and limited adaptability to dynamic wireless environments. Although Mixture-of-Experts (MoE) architectures can mitigate these issues through sparse activation, existing MoE-based semantic systems rely on inherently single-driven routing mechanisms—utilizing either source image content or channel state information (CSI) exclusively. This prevents the joint synthesis of visual semantics and fluctuating channel conditions, fundamentally limiting the overall system robustness. To address this limitation, we propose a novel multi-stage end-to-end image semantic communication system for multi-input multi-output (MIMO) channels, built upon an adaptive MoE Swin Transformer block. Specifically, we introduce a dynamic expert gating mechanism that jointly evaluates both real-time CSI and the semantic content of input image patches to compute adaptive routing probabilities. By selectively activating only a specialized subset of experts based on this joint condition, our approach breaks the rigid coupling of traditional adaptive methods and overcomes the bottlenecks of single-driven routing. Simulation results indicate a significant improvement in reconstruction quality over existing methods while maintaining the transmission efficiency.
\end{abstract}


\section{Introduction}
The rapid growth of data-intensive applications and the evolution toward next-generation wireless networks have revealed the inherent limitations of traditional communication systems \cite{6g}. Built upon Shannon's classical information theory, conventional communication paradigms primarily focus on the accurate and reliable transmission of raw bit streams, while largely overlooking the underlying meaning of the transmitted data. To overcome impending bandwidth bottlenecks and achieve more intelligent connectivity, semantic communication(SemCom) has merged as a transformative paradigm. By shifting the fundamental focus from ``how to transmit bits accurately'' to ``how to convey meaning effectively'', SemCom approach directly extracts, transmit and interprets the core semantics of information, thereby promising unprecedented improvements in transmission efficiency and system robustness\cite{semcom}.

% Driven by the remarkable breakthroughs in artificial intelligence, deep learning (DL) has established itself as the mainstream approach for realizing modern semantic communication systems. Among various DL-based solutions, the Transformer architecture has merged as the most prominent one, particularly in the realm of image semantic communication. Thanks to its powerful self-attention mechanism, a growing body of literature has adopted Transformer and its variants \cite{vit, dit} to design image semantic transceivers\cite{vit_jscc, swin_jscc}, which demonstrate exceptional capability in capturing long-range spatial dependencies and extracting global visual contextual features. Nevertheless, in the conventional dense architectures, scaling up the model to achieve better visual fidelity inevitably entails a proportional and explosive increase in both the total parameter count and the active computation overhead. During the inference phase, every single image patch must be processed by the entire network, leading to massive floating-point operations and unacceptable execution latency, which imposes prohibitive challenges for improving communication efficiency. 

The remarkable advances of deep learning (DL) has become the key enabler for SemCom. For example, in SemCom approaches for images transmission, the adopted DL architectures have evolved from relatively simple convolutional neural networks (CNNs)\cite{deep_jscc} to more sophisticated Transformer-based models\cite{vit_jscc}. However, the increased model complexity that brings performance gains also leads to substantially higher computational overhead and inference latency, thereby severely constraining communication efficiency. Furthermore, most existing approaches rely on directly trained, fixed models that lack adaptability to diverse transmission content and dynamic channel conditions, resulting in limited generalization capability and reduced robustness in practical deployments.

\begin{figure*}
    \centering
    \includegraphics[width=0.75\linewidth]{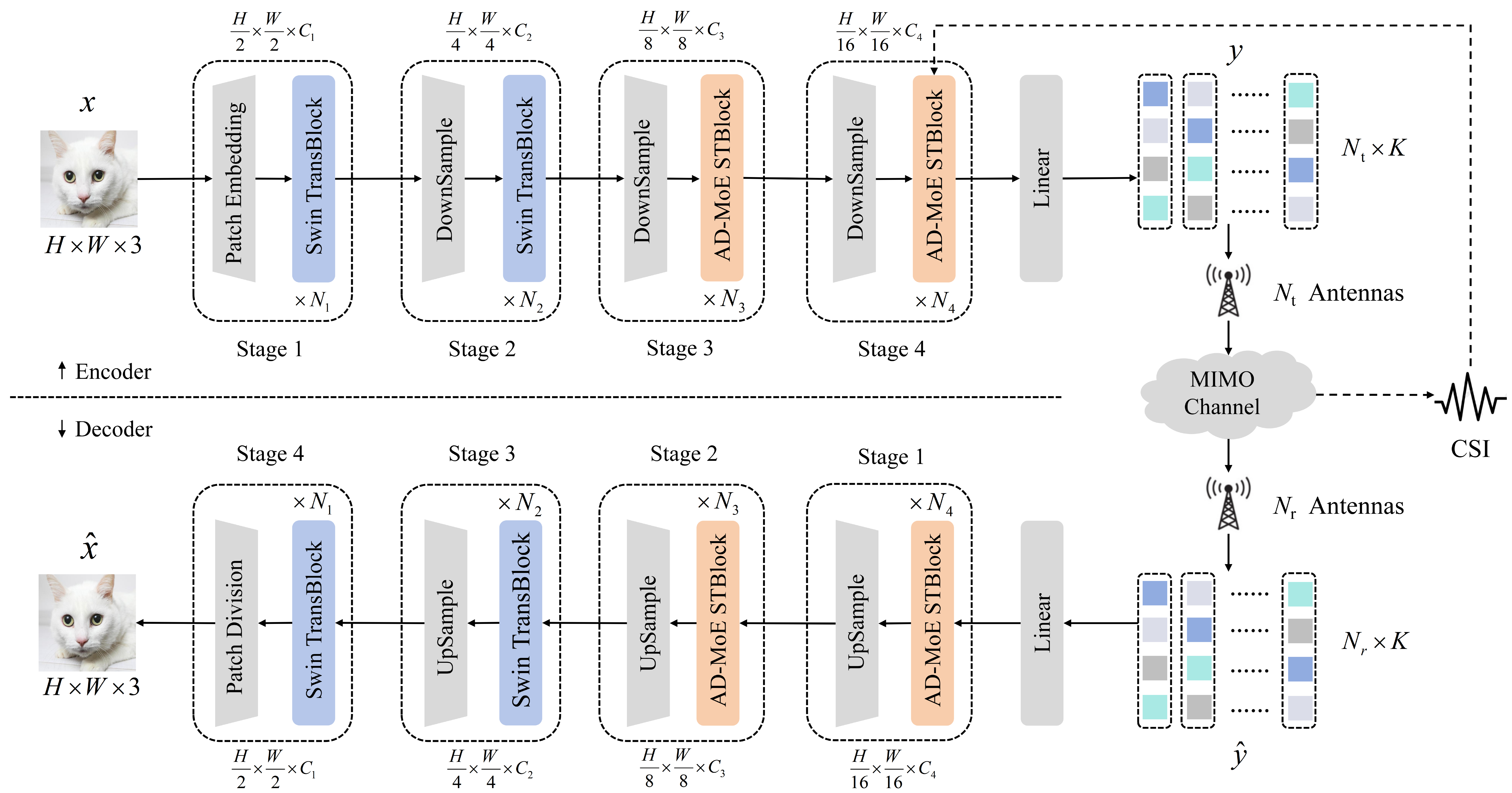}
    \caption{The overall structure of the proposed scheme for wireless image transmission.}
    \label{fig:system model}
\end{figure*}

\begin{figure*}
    \centering
    \includegraphics[width=0.75\linewidth]{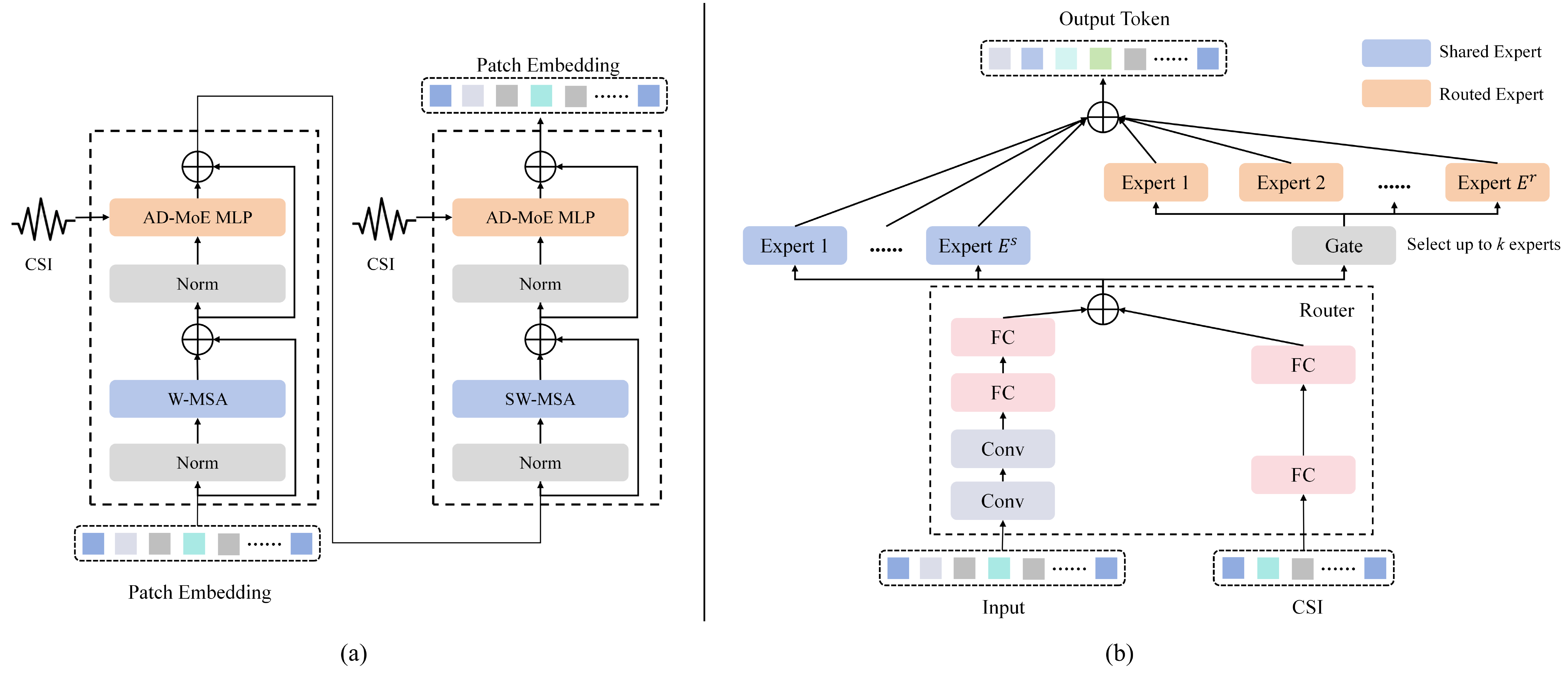}
    \caption{(a) The specific architecture of two successive AD-MoE STBlock.  (b) The architecture of AD-MoE MLP Block. $E^s$, $E^r$ denotes the number of shared experts and the number of routed experts in AD-MoE MLP blocks respectively.}
    \label{fig:ad_moe}
\end{figure*}

To address the aforementioned lack of adaptability, recent research has pivoted towards adaptive SemCom paradigms, primarily focusing on content-adaptive and channel-adaptive transmission strategies. For instance,  Fu et. al \cite{content_aware} proposed a content-aware framework to fuse global and task-related semantics via attention mechanisms. To combat dynamic wireless environment, JSCCformer-f \cite{jscc_former_f} exploits block-wise channel feedback to continually refine transmitted semantic symbols based on real-time CSI and receiver beliefs. Similarly, Zhang et. al. \cite{snr_eq_jscc} proposed a scheme called SNR-EQ-JSCC, which achieves lightweight channel adaptation by embedding the SNR directly into Transformer block and dynamically adjusting attention scores through SNR-based queries. However, these methods typically rely on auxiliary modules to process and directly fuse side information with semantic features. This rigid coupling increases computational overhead and compromises feature purity, hindering efficient scalability. To solve this problem, the MoE architecture, which is widely applied in several DL fields\cite{deepseek_moe, admv_moe}, has emerged as a highly promising solution through its sparse activation mechanism. While preliminary efforts \cite{diff_moe, moe_video} have introduced the MoE paradigm into SemCom, these existing works typically focus on either content-driven robustness or channel-driven adaptation in isolation. The absence of a unified MoE framework that jointly synthesizes visual semantics with fluctuating channel conditions fundamentally limits the system's overall adaptability, highlighting the critical need for a joint-driven routing mechanism.

To address the aforementioned limitations, we propose a novel multi-stage end-to-end image SemCom system for MIMO fading channels, empowered by an adaptive MoE Swin Transformer architecture. We design an adaptive MoE Swin Transformer block that overcomes the bottlenecks of single-driven models. It uniquely synthesizes both real-time channel state information (CSI) and the semantic content of input image patches to jointly determine the optimal routing strategy. We introduce a specialized gating mechanism within the MoE layer. By evaluating the joint condition features, it adaptively computes routing probabilities to sparsely activate only a targeted subset of experts, thereby expanding model capacity and adaptability without incurring proportional computational overhead. Building upon the proposed blocks, we develop a comprehensive multi-stage semantic transmission framework. Extensive simulation results demonstrate that our approach achieves considerable performance gains in image reconstruction quality compared to existing methods, while strictly preserving transmission efficiency.

\section{The proposed scheme}
In this section, we present the overall architecture of the proposed approach. We also introduce Adaptive MoE Swin Transformer block applied in the proposed system for image transmission.

\subsection{Overall Architecture}
An overview of the proposed architecture for wireless image transmission is given in Fig. 1. The first two stages consists of convolution layers for fast feature extraction at higher resolution, while the last two stages employ Adaptive MoE Swin Transformer blocks (AD-MoE STBlocks) to further extract images features at lower resolutions. Specifically, an RGB image $\boldsymbol{x} \in \mathbb{R}^{H \times W \times 3}$ is first split into patches with size $\frac{H}{2} \times \frac{W}{2} \times C_0$ and projected into a $C_1$ dimensional embedding space by a $5 \times 5$ CNN layer with stride of 2. After patch embedding, $N_1$ Swin Transformer blocks are applied on the latent code. The downsamplers at the beginning of the stage 2, 3 and 4 consist of a layer normalized fully connected (FC) layer which reduce the width and height of extracted latent code by half.

At the stage 2, the downsampled code is fed into $N_2$ Swin Transformer blocks. Then at the stage 3, we employ $N_3$ AD-MoE STBlocks for the low-resolution code. As shown in Fig. 2(a), AD-MoE STBlocks consist of a multi-head attention module and an adaptive MoE MLP block, in which different expert modules are applied to process input according to its feature. Similar to stage 3, this code is then sent into stage 4 which is encapsulated by a downsampling block and $N_4$ AD-MoE STBlocks. We also introduce CSI into AD-MoE STBlocks to influence the activated expert. Next, an FC layer is applied on this embedding to project it to the desired $C$ dimension. Then this projected output is allocated across the transmit antennas and time slots, forming a transmission matrix $\boldsymbol{y} \in \mathbb{C}^{N_t \times K}$, where $K$ denotes the number of channel uses dedicated to transmitting one image. We employ bandwidth ratio $R$ to represent the average number of available channel symbols per source dimension, which is defined as
\begin{equation}
    R = \frac{K}{H \times W \times 3},
\end{equation}

\begin{figure*}
    \centering
    \includegraphics[width=0.99\linewidth]{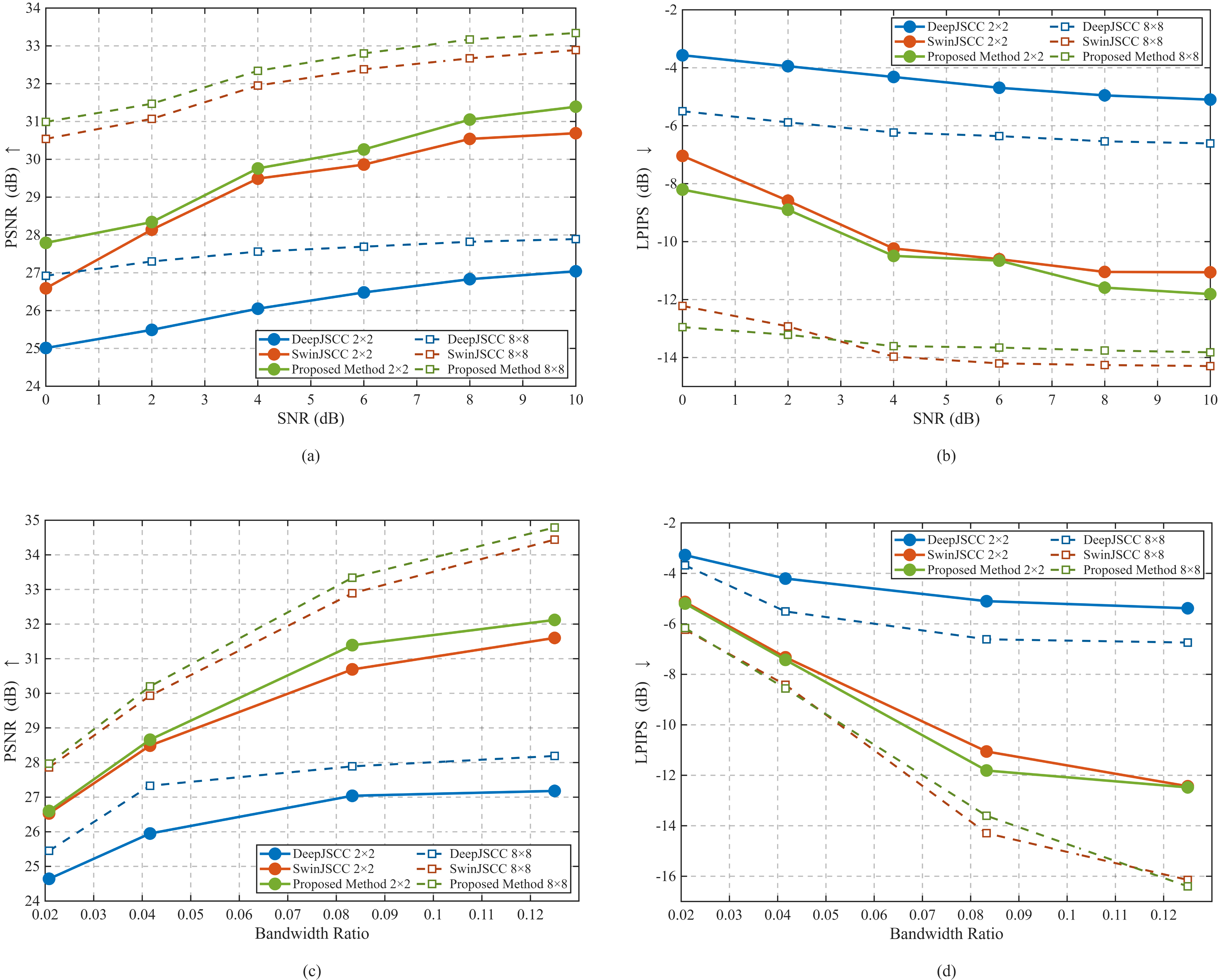}
    \caption{(a)$\sim$(b) PSNR and LPIPS performance of different models versus SNR under MIMO fading channels for the Kodak dataset, with R of 0.0833; results are shown for $2\times2$ and $8\times8$ transmit-receive antenna configurations. (c)$\sim$(d) PSNR and LPIPS performance of different models versus R under MIMO fading channel of Kodak Dataset, with SNR of 10 dB; results are shown for $2\times2$ and $8\times8$ transmit-receive antenna configurations.}
    \label{fig:SNR comparison}
\end{figure*}

\begin{figure*}
    \centering
    \includegraphics[width=0.8\linewidth]{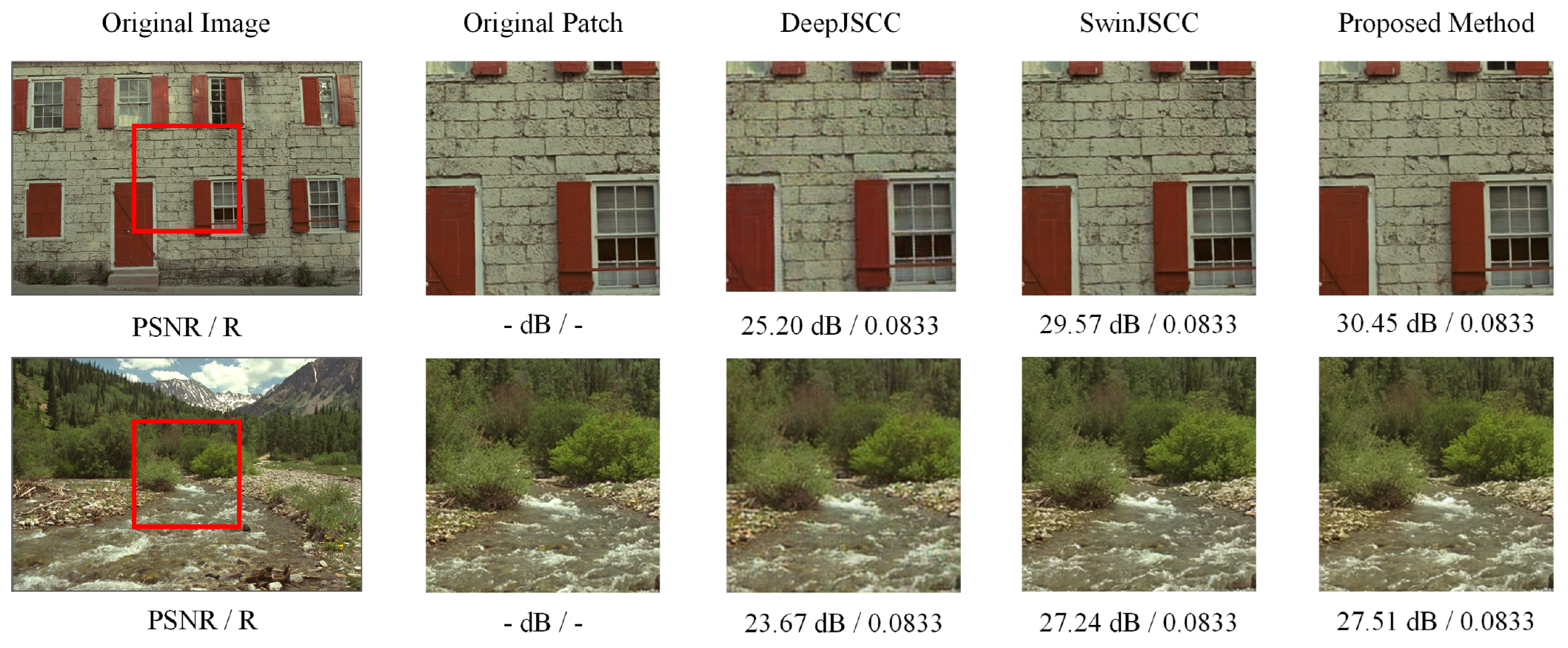}
    \caption{Examples of visual comparison under MIMO fading channel at SNR = 10dB, R = 0.0833, $N_t$ = 8, $N_r$ = 8. The first column, second column, and third to fifth column shows the original image, original patch, and reconstructions of different transmission schemes, respectively.}
    \label{fig:Example comparison}
\end{figure*}

Before transmitting the embedding $\boldsymbol{y}$ into the wireless channel, we apply the power normalization to enable $\boldsymbol{y}$ to satisfy the average power constraint. Then the embedding is directly sent into the channel. In this paper, we consider the MIMO fading channel model consisted of $N_t$ transmitting antennas and $N_r$ receiving antennas with transfer function:
\begin{equation}
    \boldsymbol{\hat{y}} = \boldsymbol{h} \boldsymbol{y} + \boldsymbol{n},
\end{equation}
where $\boldsymbol{h} \in \mathbb{C}^{N_r \times N_t}$ denotes the channel matrix, $\boldsymbol{n} \sim \mathcal{CN}\left(0, \sigma^2\right)$ denotes the Gaussian noise of the channel.

The proposed decoder has a symmetric architecture with encoder. It consists of a patch division layer, upsampling layers, AD-MoE STBlocks and a FC layer. It reconstructs input images $\boldsymbol{\hat{x}}$ from latent code $\boldsymbol{\hat{y}}$.

\subsection{Adaptive MoE Swin Transformer Block}
For the latent representation $\boldsymbol{z}_{m}$, we follow the standard Swin Transformer design for spatial feature extraction. As shown in Fig. 2(a), window-based multi-head self-attention (W-MSA) and shifted-window multi-head self-attention (SW-MSA) are applied alternately across successive transformer blocks to capture both local dependencies and cross-window interactions, which can be expressed as:
\begin{equation}
    \boldsymbol{z}^{\prime} = \text{MSA} \left( \text{Norm} \left( \boldsymbol{z}_{m} \right) \right) + \boldsymbol{z}_{m}
\end{equation}
where $\text{MSA}\left(\cdot\right)$ denotes either W-MSA or SW-MSA depending on the block.

Instead of the conventional MLP used in feed-forward layers, we introduce the proposed AD-MoE MLP blocks. As shown in Fig. 2(b), AD-MoE MLP consists of $E^s$ shared experts and $E^r$ routed experts according to the suggestions in \cite{deepseek_moe}. The shared experts are always activated to maintain training quality, while the routed experts are dynamically selected through router that jointly considers the input and CSI. Meanwhile, a gating module is introduced to dynamically determine the set of activated routed experts according to the routing weights produced by the router. Instead of activating a fixed number of experts, the proposed gating mechanism adaptively adjusts the number of active experts to better match the input characteristics and channel conditions. 

Specifically, the routed experts are first ranked according to their routing weights. The gating process starts by selecting the expert with the largest weight. Then, the weight difference between the currently selected expert and the next candidate expert is examined. If this difference is smaller than a predefined threshold $T$, the candidate expert is also activated, indicating that both experts contribute comparably to the current input. Otherwise, the selection process stops. This procedure continues sequentially until either the weight difference exceeds the threshold $T$ or the number of activated experts reaches the predefined upper bound $k$. Finally, the outputs of the shared and routed experts are aggregated to produce the FFN output. This process can be expressed as:
\begin{equation}
    \boldsymbol{z}_{t} = \text{AD-MoE MLP}\left(\text{Norm}\left(\boldsymbol{z}^{\prime}\right), \text{CSI}\right) + \boldsymbol{z}^{\prime}
\end{equation}
where $\boldsymbol{z}_{t}$ is the output of the adaptive MoE Transformer block.

\subsection{Training Loss}
During training, we employ mean square error (MSE) between reconstructed and original images as the loss function. Additionally, we adopt an additional three-component loss function designed to address the expert collapse problem. The base load balance loss ensures balanced utilization of routed experts, preventing over-activation of certain experts while others remain idle, thereby enhancing overall model efficiency. It can be expressed as:
\begin{equation}
    L_{b} = E_{r} \cdot \sum_{i=1}^{E_r} f_i \cdot p_i
\end{equation}
where $f_i$ denotes $i$th expert's activation frequency; $p_i$ represents $i$th routed expert's average weight.

To further promote routing diversity and prevent the router from converging to deterministic assignments, an entropy regularization term is incorporated, which is defined as:
\begin{equation}
    L_{e} = - \frac{1}{B} \sum_{i=1}^{B} \sum_{j=1}^{E_r} p_{ij} \cdot \log(p_{ij} + \epsilon)
\end{equation}
where $B$ denotes the training batch size, $p_{ij}$ is the $j$th expert's activation frequency of $i$th batch, $\epsilon$ denotes a small constant for numerical stability. 

The third component penalize high variance in expert utilization rates to prevent the scenarios where certain experts are systematically underutilized, and it can be expressed as:
\begin{equation}
    L_v = E_r \cdot Var\left( \boldsymbol{f} \right) =  \sum_{i=1}^{E_r} \left( f_i - \frac{1}{E_r}\sum_{j=1}^{E_r}f_j \right)^2
\end{equation}
where $\boldsymbol{f}$ is the expert's activation frequency vector.

% The total loading balance loss aggregates the three components into a unified objective:
% \begin{equation}
%     L_{MoE} = L_b + \alpha L_e + \beta L_v
% \end{equation}
% where $\alpha$ and $\beta$ are hyperparameters controlling $L_{e}$ and $L_{v}$ respectively.

The total training loss is defined as the weighted sum of these two loss items, denoted by:
\begin{equation}
    L = L_{MSE} + \lambda L_b- \alpha L_e + \beta L_v
\end{equation}
where $\lambda$, $\alpha$ and $\beta$ are hyperparameters controlling loss.

% The load balance loss ensures balanced utilization of routed experts, preventing over-activation of certain experts while others remain idle, thereby enhancing overall model efficiency. It can be expressed as:
% \begin{equation}
%     L_{b} = \frac{1}{N_{MoE}} \sum_{i=1}^{N_{MoE}} \boldsymbol{f_i}p_i
% \end{equation}
% where $N_{MoE}$ is the number of AD-MoE modules; $\boldsymbol{f_i}$ denotes each expert's activation frequency vector of $i$th AD-MOE module; $\boldsymbol{p_i}$ represents each routed expert's average weight vector of $i$th AD-MOE module.

% The gate loss dynamically adjusts the number of activated experts per forward pass, reducing computational overhead while maintaining performance to improve inference efficiency. It is formulated as:
% \begin{equation}
%     L_{gate} = \frac{1}{N_{MoE}} \sum_{i=1}^{N_{MoE}} \frac{k_i^{ad}}{k_i},
% \end{equation}
% where $k_i^{ad}$ denotes the mean activated experts number of $i$th AD-MOE module; $k_i$ denotes the maximum activated number of $i$th AD-MOE module.

\section{Experimental Results}

\subsection{Experimental Setup}
\noindent \textbf{Datasets:} we train and evaluate the proposed scheme on image dataset: DIV2K \cite{div2k}. DIV2K contains 900 high-resolution images for training. We evaluate the model trained on the trainset of the DIV2K dataset by Kodak dataset \cite{kodak}. During both training and testing, images from Kodak datasets are cropped into the resolution of $256 \times 256$.

\noindent \textbf{Benchmarks:} We compare the proposed scheme with DeepJSCC \cite{deep_jscc} and SwinJSCC\cite{swin_jscc}. It should be noted that the SwinJSCC adopted here is not equipped with SNR adaptation modules or compression ratio adaptation modules, aiming to demonstrate the fundamental performance of Swin Transformer in image transmission.

% Specifically, we employ the BPG \cite{bpg} code for compression combined with 5G LDPC \cite{ldpc} codes for channel coding (marked as "BPG + LDPC"). For "BPG + LDPC" scheme, we use 5G LDPC with code length of 6144, and choose the best-performing configuration of coding rate and modulation under each specific SNR.

\noindent \textbf{Metrics:} We qualify our system's performance in terms of image reconstruction quality and bandwidth efficiency. To access image reconstruction quality, we use two widely adopted metrics: Peak Signal-to-Noise Ratio (PSNR) and Learned Perceptual Image Patch Similarity (LPIPS) \cite{lpips}. To ensure readability, we convert LPIPS into an exponential representation, namely $LPIPS\left(dB\right)=10\log_{10}\left(LPIPS\right)$.

% PSNR measures the pixel-level difference between two images, denoted by
% \begin{equation}
%     PSNR=10 \times \log_{10}{\frac{Max(\boldsymbol{x},\boldsymbol{\hat{x}})^2}{MSE(\boldsymbol{x},\boldsymbol{\hat{x}})}} ,
% \end{equation}
% where $Max(\boldsymbol{x},\boldsymbol{\hat{x}})^2$ represents the maximum difference between pixel values in the images. LPIPS measures the distance in the feature space derived by the pre-trained AlexNet from a human perceptual perspective, denoted by
% \begin{equation}
%     LPIPS(\boldsymbol{x},\boldsymbol{\hat{x}})=\sum\limits_{l}\frac{1}{H_lW_l}\sum\limits_{i,j} \left\Vert c_l\odot \left(f_{ij}^l-\hat{f_{ij}^l}\right)\right\Vert_2^2 ,
% \end{equation}
% where $f^l_{i,j}$ and $\hat{f^l}_{i,j}$ denote the $\left(i,j\right)$th element in normalized latent feature maps output by layer $l$ of the AlexNet, respectively. $H_l$ and $W_l$ denote the height and width of the feature map, respectively. Notation $\odot$ denotes the scale operation and $c_l$ represents the pretrained weights for intermediate latent output by layer $l$. 

\noindent \textbf{Training Details:} We use 4 stages with $\left[ N_1, N_2, N_3, N_4 \right]=[2,2, 6, 2]$, $\left[ C_1, C_2, C_3, C_4 \right]=[128, 192, 256, 320]$, $E^s=1$, $E^r=5$, $k=2$, and window size is set to 8. All models are trained until convergence on NVIDIA RTX A6000 GPUs, with the initial learning rate of $10^{-4}$. 

\subsection{Results Analysis}
% Fig. 3(a)$\sim$3(b) show the PSNR performance versus SNR over the AWGN channel, and Fig. 3(c)$\sim$3(d) present the LPIPS performance versus SNR over AWGN channel. Compared to the CNN-based DeepJSCC scheme, we achieve much better performance for all SNRs. For the FFHQ $64\times64$ dataset, the proposed method significantly outperform the DeepJSCC scheme and the "BPG + LDPC" in both PSNR and LPIPS comparison. For high-resolution images in Kodak dataset, our proposed method consistently outperforms DeepJSCC scheme across both PSNR and LPIPS metrics. Furthermore, under low SNR conditions, our approach also surpasses the "BPG + LDPC" scheme in terms of both PSNR and LPIPS performance. 

Fig. 3(a)$\sim$3(b) show the PSNR and LPIPS performance versus SNR over MIMO fading channel, and Fig. 3(c)$\sim$3(d) present the PSNR and LPIPS performance versus R over MIMO fading channel. For images in Kodak dataset, the proposed method significantly outperform the DeepJSCC scheme in both PSNR and LPIPS comparison of different MIMO modes. Compared to the SwinJSCC scheme, our approach also achieve better PSNR performance for all SNRs and Bandwidth ratios. As for LPIPS performance, our proposed method still surpasses the SwinJSCC scheme for most SNRs and Bandwidth ratios. The comparison between our proposed method and SwinJSCC can also be regarded as an ablation study. Clearly, with the addition of AD-MoE STBlock, the model's performance in image transmission has been significantly improved. These findings conclusively validate the effectiveness of our proposed AD-MoE strategy. Fig. 4 illustrates examples of reconstructed images produced by DeepJSCC, SwinJSCC and our proposed method, corroborating our results directly.

% Fig. 3(c)$\sim$3(d) present the PSNR and LPIPS performance versus R over MIMO fading channel. Our proposed model can generally outperform DeepJSCC and SwinJSCC for most bandwidth ratios. The comparison result between our proposed method and SwinJSCC also prove the considerable gains of the AD-MoE module. These results demonstrates that our proposed model maintains consistent performance under various bandwidth ratios. Fig. 4 illustrates examples of reconstructed images produced by DeepJSCC, SwinJSCC and our proposed method, corroborating our results directly.

\begin{figure}
    \centering
    \includegraphics[width=0.7\linewidth]{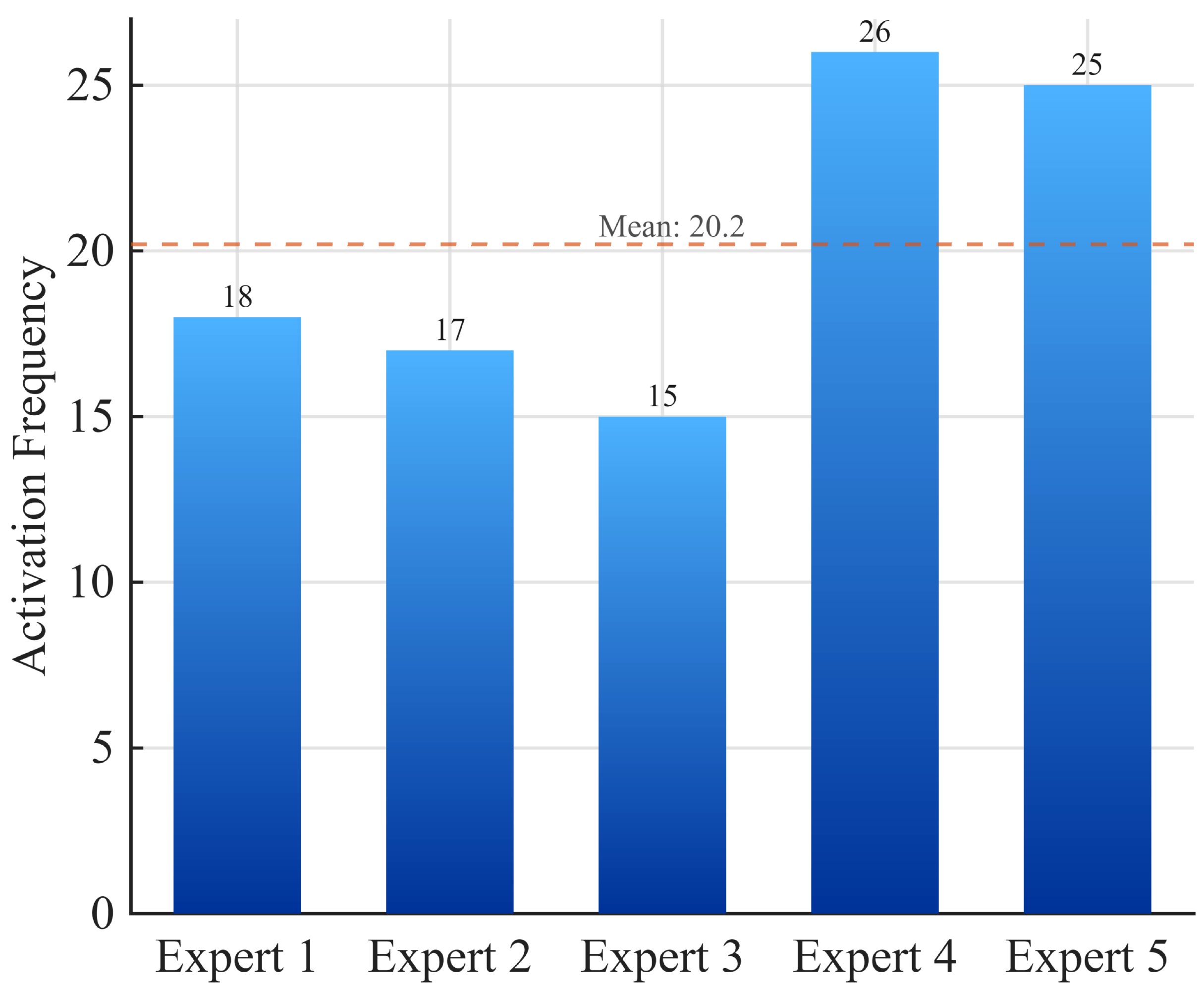}
    \caption{Expert activation frequency in the last AD-MoE STBlock of the final encoder layer, evaluated on the CLIC2021 dataset.}
    \label{fig:cr}
\end{figure}

To visually demonstrate the effect of expert activation in the AD-MoE STBlock, we tested the model on 60 images from the CLIC2021 dataset\cite{clic} under conditions of SNR = 10 dB and R = 0.0833, and recorded the expert activation status in the last layer of the encoder, as shown in Fig. 5. Guided by the routing mechanism, the model activated the 5 experts in the AD-MoE STBlock in a relatively balanced manner, ensuring that each expert was sufficiently utilized and avoiding the overuse of specific experts. In this AD-MoE STBlock, an average of 1.683 experts were activated per image during transmission, demonstrating the model's ability to dynamically control the number of activated experts.

% Meanwhile, in Fig. 3(a)$\sim$3(d), we also present ablation studies to compare performance with and without the AD-MoE scheme. We revert the AD-MoE MLP blocks in Transformer blocks to single MLP blocks to validate the performance of systems excluding AD-MoE scheme. As shown in results, the system employing the AD-MoE strategy achieves superior performance compared to its counterpart without this module on the low-resolution FFHQ $64\times64$ dataset. For high-resolution images in Kodak dataset, the system with AD-MoE Transformer blocks demonstrates even more pronounced advantages in both PSNR and LPIPS metrics. These findings conclusively validate the effectiveness of our proposed AD-MoE strategy. Fig. 4 illustrates examples of reconstructed images produced by DeepJSCC, "BPG + LDPC" and our proposed method, corroborating our results directly.

% Fig. 5 present the PSNR performance versus the channel bandwidth ratio over the AWGN channel of Kodak dataset. Our proposed model can generally outperform DeepJSCC and "BPG + LDPC" for most CBRs. Meanwhile, our model achieves considerable gains compared to model without the AD-MoE module. These results demonstrates that our proposed model maintains consistent performance under various channel bandwidth ratios.

\section{Conclusion}
In this paper, we proposed a highly efficient and robust SemCom system for wireless image transmission by integrating a novel Adaptive MoE architecture into the Swin Transformer backbone. To overcome the limitations of conventional expert routing, we designed an innovative Adaptive MoE Swin Transformer block that jointly synthesizes input visual semantics and instantaneous MIMO channel state information to formulate the optimal routing strategy. Coupled with a dynamic expert gating mechanism, our system intelligently modulates the number of activated experts for, successfully breaking the rigid trade-off between model capacity and computational complexity. Comprehensive evaluations on standard image datasets confirm that the proposed scheme significantly outperforms existing methods in terms of both reconstruction fidelity and bandwidth efficiency.

% This work proposed a novel semantic communication system leveraging MoE architecture for wireless image transmission. We employed Swin Transformer blocks whose MLP blocks in feed-forward layer are replaced with MoE structure for image feature extraction and reconstruction. Additionally, we introduce an input adaptive expert activation mechanism, which enables the router to dynamically modulate the number of activated experts based on input token, thereby optimizing resource allocation while preserving task performance. Then we trained our model on popular datasets and evaluated its transmission performance. Experimental results validate a remarkable improvement on reconstruction quality and transmission efficiency over existing methods.

\bibliographystyle{IEEEtran}
\bibliography{reference}

\end{document}